%% file: mmd.tex
\documentclass[10pt,twocolumn,letterpaper]{article}

\usepackage{wacv}
\usepackage{booktabs} 
\usepackage{amsmath}       
\usepackage{breqn}
\usepackage{amssymb}
\usepackage[linesnumbered,boxed,commentsnumbered, ruled]{algorithm2e}
\usepackage{algpseudocode}
\usepackage{graphicx}
\graphicspath{{images/}}

\usepackage{times}
\usepackage{epsfig}
\usepackage{subfigure}
\usepackage[mathscr]{euscript}
\usepackage{verbatim,color}

\wacvfinalcopy 

\ifwacvfinal\fi
\begin{document}

\title{Multi-modal dialog for browsing large visual catalogs using exploration-exploitation paradigm in a joint embedding space}

\author{Indrani Bhattacharya \thanks{These authors contributed equally to this work.} \\
{\small Rensselaer Polytechnic Institute}\\
{\tt\small bhatti@rpi.edu}
\and
Arkabandhu Chowdhury \footnotemark[1]\\
{\small Rice University}\\
{\tt\small arkabandhu@rice.edu}
\and
Vikas Raykar \\
{\small IBM Research, India}\\
{\tt\small viraykar@in.ibm.com }
}

\maketitle

\ifwacvfinal\thispagestyle{empty}\fi

\input{abs}
\input{introduction}
\input{review}
\input{data_creation}
\input{method}

\input{results}

\input{conclusion}

{\small
\bibliographystyle{ieee}
\bibliography{mmd}
}

\end{document}

%% file: abs.tex
\begin{abstract}
We present a multi-modal dialog system to assist online shoppers in visually browsing through large catalogs. Visual browsing is different from visual search in that it allows the user to explore the wide range of products in a catalog, beyond the exact search matches. We focus on a slightly asymmetric version of the complete multi-modal dialog where the system can understand both text and image queries, but responds only in images. We formulate our problem of ``showing $k$ best images to a user'' based on the dialog context so far, as sampling from a Gaussian Mixture Model in a high dimensional joint multi-modal embedding space, that embed both the text and the image queries. Our system remembers the context of the dialog and uses an exploration-exploitation paradigm to assist in visual browsing. We train and evaluate the system on a multi-modal dialog dataset that we generate from large catalog data. Our experiments are promising and show that the agent is capable of learning and can display relevant results with an average cosine similarity of 0.85 to the ground truth. Our preliminary human evaluation also corroborates the fact that such a multi-modal dialog system for visual browsing is well-received and is capable of engaging human users.
\end{abstract}

%% file: introduction.tex
\section{Introduction}\label{sec:intro}
When a customer visits a physical retail store, she has the option to quickly \textbf{visually browse} through a large number of products. In most cases, she has the option to engage in a one-on-one conversation with a human agent in the store who can assist the customer to find what she is looking for. For example, the customer may say ``Show me blue dresses'' and the agent can respond by displaying blue dresses. If the customer points to a particular dress and says ``Show me more like this'', the agent not only shows products that are exactly similar to the chosen product, but also shows a few samples that he thinks the customer may like, allowing the customer to explore the catalog better and refine her preferences. The human agent understands and remembers the customer's requirements, learns her likes and dislikes and pro-actively makes recommendations in an \textbf{exploration-exploitation} framework, which in turn allows the customer to know about the range of products in the store quickly and also refine her preferences efficiently. 

\begin{figure}[hbtp!]
  \centering
  \includegraphics[trim={2.6cm 8.2cm 8.2cm 2.5cm},clip, width=0.7\columnwidth]{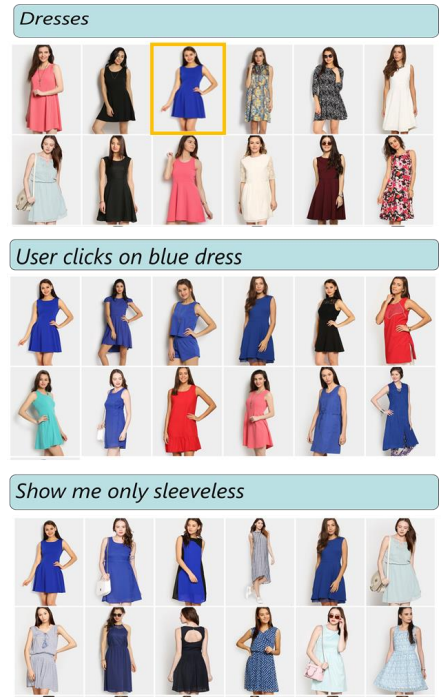}
  \caption{Visual browsing through multi-modal dialog.}
  \label{fig:dialog}
\end{figure}
\vspace{-3mm}
In contrast, an online customer does not have much flexibility to  quickly visually browse through large catalogs in an exploration-exploitation setup. More often than not, a customer rarely goes beyond the first 2-3 pages of search results returned by the search engine. Also, search engines mostly restrict search results to exact matches to the search query, without any exploration. Since online catalogs are huge there is a significant fraction of products which never surface on the the first page of the search results unless the user is specifically searching for it or it is being promoted by the retailer. 
In this paper, we present a system to assist customers in visually browsing through large catalogs in an exploration-exploitation setup using multi-modal dialog that allows the customer to search better,  refine her preferences and at the same time ensure that the customer gets to see a large part of the catalog. Figure \ref{fig:dialog} shows an example of our proposed system. The key aspects of our system are: \\
\begin{itemize}
\item \textbf{Visual browsing is different from visual search:} Online visual browsing is different from visual search in that it allows the user to explore the catalog beyond the exact search results, thus capturing the concept of exploration-exploitation. When an online customer expresses her preference in the form of an image click, it is difficult to exactly verbalize what the user wants. For example in the second round of the dialog given in Figure \ref{fig:dialog}, when the customer clicks on the third blue dress in the first row, does the customer want dresses similar in design to the one clicked? Or does the customer want only blue dresses, irrespective of the design? Is the customer willing to explore tops that are not blue? In this example, the agent not only shows blue dresses in response to the user click, but also explores and shows a red long top, a blue top and a short, and a couple of dresses that are not blue, but similar to the clicked dress in design. The advantages of visual browsing are: (1) it allows for easier navigation through large product catalogs, and (2) it provides the user the capability to sample from the whole catalog based on the exploration-exploitation paradigm. Such a system allows for a more interactive online shopping experience, while providing an opportunity to quickly browse through representative products of the catalog.\\
\item \textbf{Multi-modal dialog:} The user can express his/her preference or query in natural language text or image click, while the agent responds only in images. This is a slightly asymmetric version of a complete multi-modal dialog. In order for the agent to respond to both text and image queries, it needs to learn a joint embedding space that maps both types of queries to a common space.\\
\item \textbf{Context:} The agent has to understand not only the current query, but also remember previous rounds of the dialog. For example in the third round of dialog in Figure \ref{fig:dialog}, although the user query is ``Show me only sleeveless", the agent remembers that the user had queried for dresses in the first round and had clicked on a blue dress in the second round of the dialog and hence only displays sleeveless dresses in various shades of blue.
\end{itemize}

The first challenge in building such an intelligent and interactive system is the absence of any existing multi-modal dataset. In this paper, we first build a multi-modal dialog dataset from large product catalogs, as described in Section \ref{sec:dataset}. We then propose the architecture of a model that can learn how to respond to multi-modal queries based on user context and the exploration-exploitation paradigm, as described in Section \ref{sec:method}. We train our agent and evaluate its performance on our multi-modal dialog dataset with promising results.  

%% file: review.tex
\section{Related Work}

Several recent works have focused on the intersection of computer vision and natural language, in the development of conversational agents, two-player games, visual question-answering, image captioning etc. Some relevant works include \cite{das2017visual}, where the authors train an AI agent to hold meaningful conversation with a human about an image. Given an image, a dialog history and a question, the agent comes up with a meaningful reply in natural language. In \cite{de2017guesswhat}, H. de Vries et al.~introduced the two-player ``Guess What?!'' game, where the task is to guess a particular object in an image. Das et al.~\cite{das2017learning} use deep reinforcement learning to develop a co-operative image guessing game between two AI agents, the Q-Bot and the A-Bot. The goal for both the bots is to steer the dialog in a way that leads the Q-Bot to quickly guess the image that is being referred to from the line-up of images.

While research in multi-modal dialog is gaining traction, the domain of online shopping is yet to tap into its full potential. E-commerce leaders like Amazon and Netflix extensively use recommender systems. Starting with content based and collaborative filtering, recommender systems are now focusing on matrix factorization, multi-armed bandits, and
methods for blending recommendations \cite{karypis2018recent}. For online shopping, visual search is very common and recent advances in computer vision have made significant contributions to this field \cite{hadi2015buy, jing2015visual, manandhar2018brand, yang2017visual}. Going beyond visual search, Laenen et al.~\cite{laenen2018web} present a multi-modal product retrieval method that leverages both text and image queries. They present a model that learns the cross-modal representations and ranks images based on their relevance to a given query. Teo et al.~\cite{teo2016adaptive} proposed a system of visual browsing on e-commerce platforms using image clicks by adaptive personalization and incorporating concepts of exploration-exploitation. They used a Bayesian regression model, a sub-modular diversification framework and personalized category preferences. 

In contrast to existing literature for online product recommendation and visual search/ visual browsing, we propose a framework of multi-modal dialog to assist online customers in visual browsing through product catalogs. The system would be capable of remembering context, correlating text and image queries to a common representation space and go beyond ``exact matches" to allow for the user to explore the catalog quickly. To the best of our knowledge, there is no existing work that unifies all these concepts to provide a one-stop intelligent agent that can mimic an in-store retail experience.

%% file: data_creation.tex
\section{Dataset Creation} \label{sec:dataset}

To the best of our knowledge, there is no existing dataset of multi-modal dialog for visual browsing of large catalogs. A primary reason is that online visual browsing is still not deployed in large scale and still not widely used on e-commerce websites. It is also difficult to collect real-world multi-modal data that mimics in-store retail experience in an on-line shopping framework. 
\begin{figure}[htbp!]
  \centering
  \includegraphics[width=0.9\columnwidth]{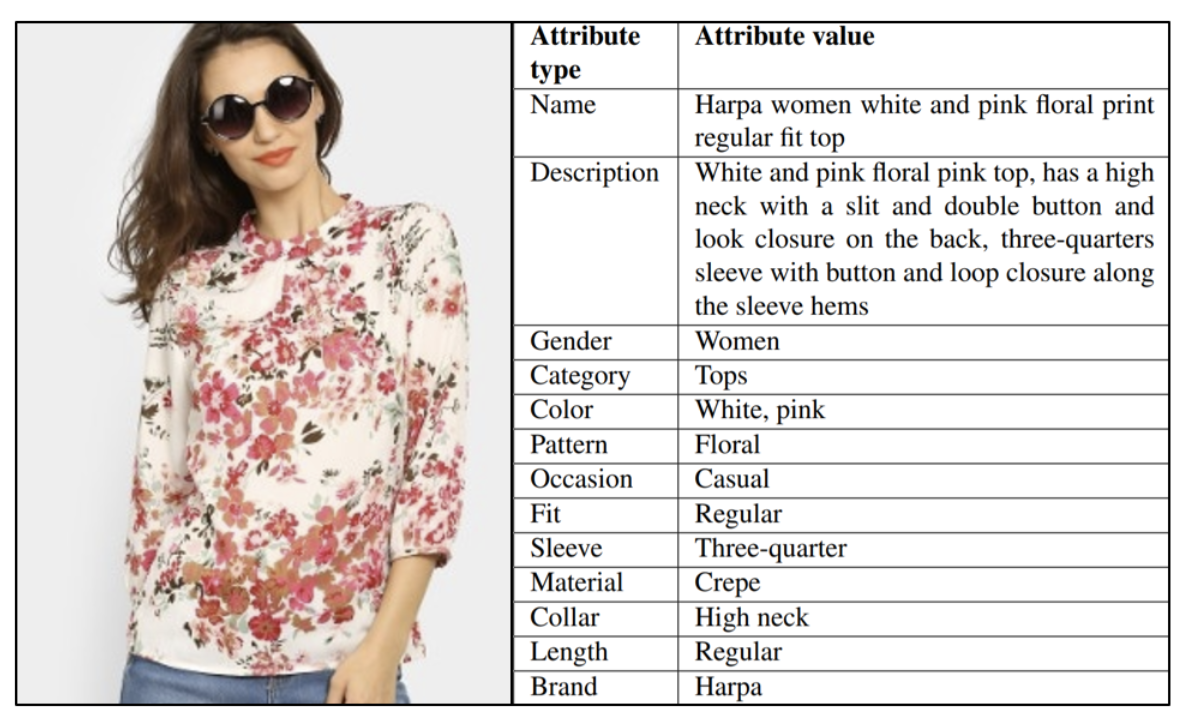}
  \caption{The image of a typical product in the catalog and some associated attributes and attribute values.}
  \label{fig:product1}
\end{figure}

In order to create the dataset, we crawled 71905 unique products (consisting of apparel, footwear, and accessories) from a fashion e-commerce site. For each product, we crawled the main product image together with available meta-data including category, name, description and various attribute values. Figure \ref{fig:product1} shoes a typical product in the catalog, with the image and some of the available meta-data information.

Since there can be a multitude of product category types, attributes and attribute values, we standardize them by building a fashion specific vocabulary. Each category has certain attributes and each attribute can take some specified set of values. Table \ref{tab:vocab} shows a snippet of our vocabulary and some sample tokens of categories (e.g. \textit{shoes}), attribute names (e.g. \textit{color}) and possible attribute values (e.g. \textit{red, blue, etc.}). Some attributes are common to all categories (e.g. \textit{color}), while some attributes are specific to only certain categories (e.g. attribute \textit{sleeves} is applicable for categories \textit{tops, shirts, dresses} and not for category \textit{shoes}). The data has a total of 130 categories, 17 attribute types, and 501 unique attribute values, resulting in a vocabulary size of 648.
\begin{table}[htbp]
\centering
\caption{Fashion specific vocabulary built by standardizing product categories, attributes and attribute values.}
\begin{tabular}{|p{0.15\columnwidth}|p{0.1\columnwidth}|p{0.6\columnwidth}|} \hline
\textbf{Attribute} & \textbf{Count}&\textbf{Sample values}\\ \hline
Gender & 2 & men, women\\ \hline
Category & 130 & shoes, dresses, trousers, shirts ... \\ \hline
Color & 46 & sky blue, peach, red, violet ... \\ \hline
Material & 65 & leather, cotton, jute, silk ...\\ \hline
Pattern & 17 & woven design, floral, embellished, checkered ...\\ \hline
Brand & 160 & John Players, Reebok, 109F, Adidas ...\\ \hline
\end{tabular}
\label{tab:vocab}
\end{table}

\begin{figure}[htbp]
  \centering
  \includegraphics[width=0.85\columnwidth]{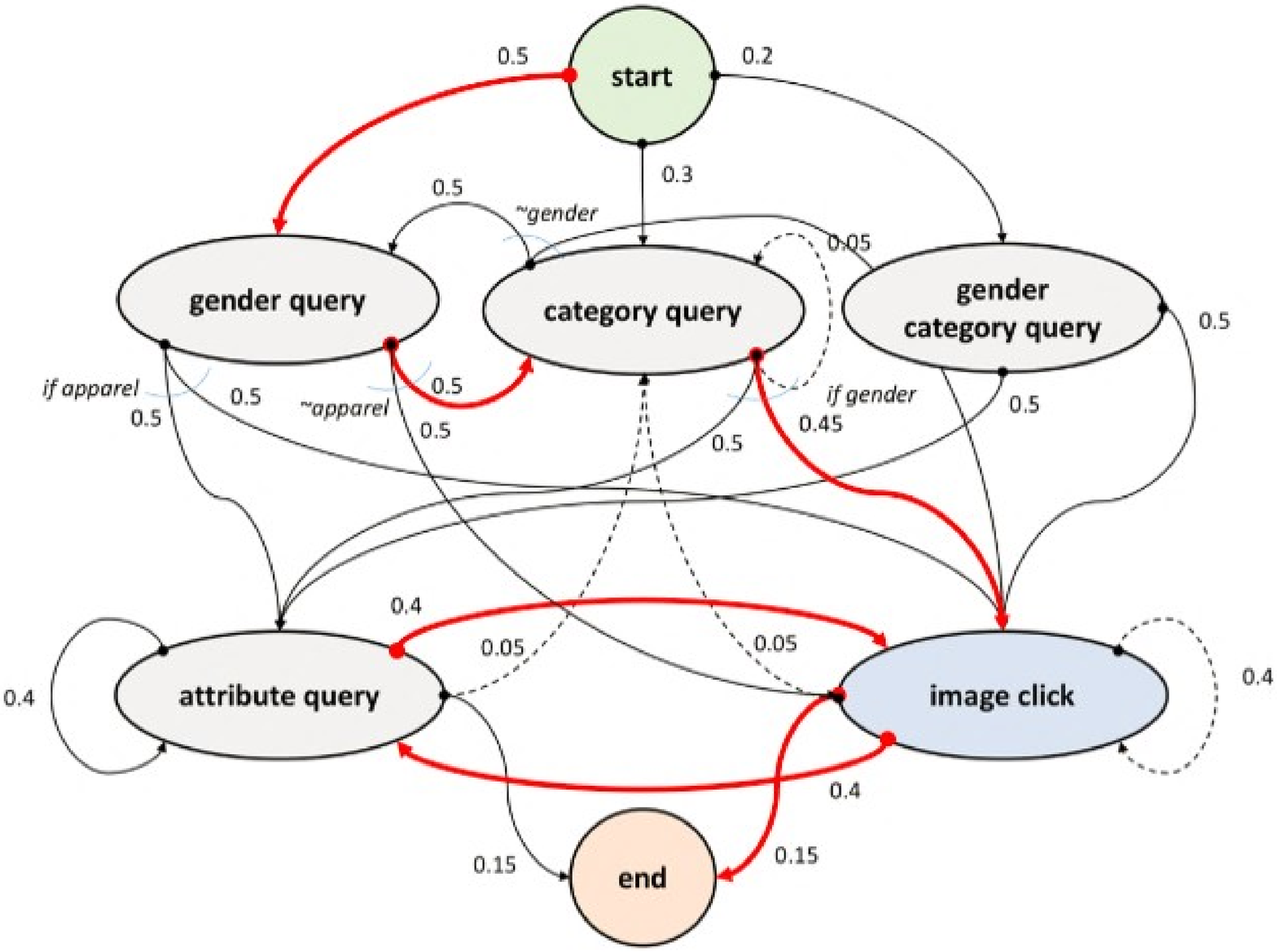}
  \caption{Multi-modal dialog dataset creation by probabilistic finite state automaton.}
  \label{fig:automaton}
\end{figure}

In our initial experiments, we select a subset of the entire product catalog, focusing only on categories related to men and women footwear, giving a total product count of 3500. We generate around 5K multi-modal dialog sessions on this dataset by a context dependent probabilistic finite state automaton, as illustrated in Figure \ref{fig:automaton}. A sample dialog starts at the \textit{start} node and continues by a random walk through the different nodes in the graph until it hits the \textit{end} node. The edge weights denote the various transition probabilities between the nodes.

To get the dialog started, at the \textit{start} node,  we transition to one of the three nodes -- \textit{gender query}, \textit{category query}, \textit{gender category query}. It may be noted that although the user can express her choices both in terms of text or image clicks, the starting query is always a text query, which is generated by randomly sampling a term from the fashion vocabulary. The \textit{gender query} node generates queries only related to gender, while the \textit{category query} node generates queries related to different categories (e.g. running shoes, formal shoes, sandals, etc.) and the \textit{gender category query} generates queries about gender specific categories (e.g. men's running shoes). The products with their attributes are indexed into an elastic search \cite{elastic}, to facilitate their retrieving from the catalog. The images corresponding to the top 6 results are shown to the user. 

From \textit{gender query}, \textit{category query} or \textit{gender category query} node, the dialog can transition to either the \textit{attribute query} node, or the \textit{image click} node. The \textit{attribute query} node generates text queries related to specific attributes. At every round of the dialog, we maintain the context, in terms of gender, category name and attributes queried. Based on the context and the current search query, we retrieve results to display to the user. The \textit{image click} node corresponds to the user clicking on any one of the displayed images in the current state of the dialog. An image-click query is simulated as a random selection of one of the 6 images shown in the previous round of the dialog. The `exploration-exploitation' paradigm is implemented in response to an image click query. The response to the image click query should include (a) some results which are very similar to the clicked image in terms of image features, and (b) some results which are somewhat different but only to the extent where the user may be interested, pertaining to the current query and the context.

\begin{figure}[htbp]
  \centering
  \includegraphics[width=0.8\columnwidth]{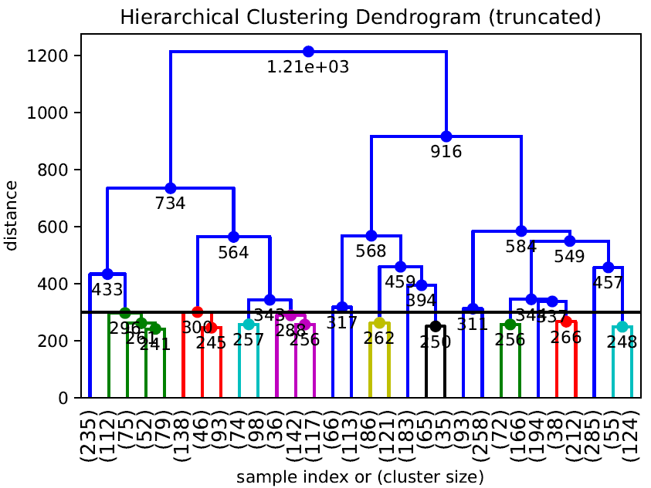}
  \caption{Hierarchical clustering to retrieve results corresponding to image-click query, by taking into account the exploration exploitation paradigm. }
  \label{fig:clust}
\end{figure}

In order to extract image features, we use the pre-trained VGG16 \cite{simonyan2014very} architecture and use the FC7 layer to represent the image features. We use a probabilistically generated number $n_1$, where $1 \leq n_1 \leq 6$, to display the $n_1$ K-Nearest Neighbors (KNN) results and $6-n_1$ results based on hierarchical clustering, as shown in Figure \ref{fig:clust}. We choose hierarchical clustering in order to capture the hierarchical  structure of products in a fashion catalog. The KNN results are meant to exploit the catalog for closest image results and the  clustering results are meant to explore the catalog. In the hierarchical clustering, the cluster size $s$ is chosen probabilistically, such that $s$ is $2 - 5$ times the maximum distance in the KNN results. This value is experimentally seen to give an optimum clustering to explore the catalog without drastically deviating from the clicked image. As the number of dialog rounds increases, $n_1$ increases, thereby focusing more and more on the user preference, with less exploration.

From the \textit{attribute query} node and the \textit{image click} query node, the dialog can proceed to the \textit{end} node with certain probability. 

In generating the simulated dialog dataset, we particularly focus on the following features:\\
\textbf{Context dependent state transition:} The transition from one node to another depends on context. For example, if we are in the \textit{gender} query node and the \textit{category} name is not yet available in the context, then we try to get the category in the current context by transitioning to the \textit{category query} node or to the \textit{image click} node.\\
\textbf{Context switches:} We allow for context switches in the dialog. This is a more realistic scenario in which the user is free to switch context. Context switches are shown as dotted lines in Figure \ref{fig:automaton}. Any time there is a context switch, the gender and category in the current context is reset to the new values.\\
\textbf{Exploitation-exploration:} We explicitly include exploration in the image-click query mode, as described above. The rationale behind this is that fashion shopping is highly serendipitous and we leave some room to the user to explore other parts of the catalog.
\vspace{-2mm}

%% file: method.tex
\section{Proposed Method} \label{sec:method}
An intelligent agent should be capable of  assisting customers without having real-time access to the meta-data information in the product catalog. It should be capable of responding to unstructured or natural language conversations or even images provided by the user not present in the catalog. Although our current set-up only allows for the user to express her preference from the existing product catalog, our vision is to extend the capabilities of the agent to respond to images that are not present in the catalog and fed in externally by the user. Thus, it is essential for the agent to comprehend multi-modal queries and learn to sample judiciously from large catalogs to assist the customers.
 
The first step for the agent in achieving this objective is to learn a common representation of images and their text descriptions in a high-dimensional joint embedding space. We use the Correlational Neural Network (CorrNet) \cite{chandar2016correlational} (described in Section \ref{sec:corrNet}) and pre-train the model with our catalog data. Once the agent learns to map an input query onto this high dimensional vector space, it has to learn how to respond to the query by showing relevant products. We model this behavior of the agent as sampling from this joint embedding space by a Gaussian Mixture Model (GMM). However, sampling is a non-differentiable operation, and cannot be used in a neural network based model that uses back-propagation. We navigate this problem by modeling the sampling operation using some re-parameterization techniques, as discussed in Section \ref{sec:sampling}. The following sections describe the complete model in more detail.

\subsection{Query encoding:} \label{sec:encode}
For image encoding, we use the pre-trained VGGNet-16~\cite{simonyan2014very} Convolutional Neural Network. The 4096 dimensional representation obtained from the FC7 layer of VGGNet-16 form our encoded image representation $\mathcal{I}_e$. Each word in the text query is encoded using the 300 dimensional representation obtained from Word2Vec~\cite{mikolov2013efficient}. 
The entire text query containing multiple words is then represented using the Continuous Bag of Words (CBOW) model \cite{mikolov2013efficient}, where we average the Word2Vec encoding over all the words in the query, and this representation forms our encoded text representation $\mathcal{T}_e$. The text/image query is then passed on to the pre-trained CorrNet model to be encoded into the joint embedding space. 

\subsection{Common Representation Learning using CorrNet} \label{sec:corrNet}
The Correlational Neural Network architecture consists of an input layer, a hidden layer, and an output layer. We pre-train this network using all the products in the catalog. One view of the input  $\textbf{Z}$ is the encoded frontal image representation $\mathcal{I}_e$ and the other view is the encoded text description $\mathcal{T}_e$ of the product formed from the attributes \textit{color, gender} and all \textit{synonymous category names}. We choose these attributes among all the available meta-data information because these features are the most representative in describing a product.

The hidden layer computes an encoded representation of dimension $k$, from the input $\textbf{Z}$ as below:
\begin{dmath}
H(Z) = f(W \mathcal{I}_e + V \mathcal{T}_e  + b)
\end{dmath}
where $W \in \mathbb{R}^{k \times 4096}$, $V \in \mathbb{R}^{k \times 300}$ and $b \in \mathbb{R}^{k \times 1}$ . The function $f$ can be any non-linear activation function, like \textit{sigmoid} or \textit{tanh}. We use the sigmoid activation function here. In our experiments, we consider the hidden layer dimension $k=200$. The output layer reconstructs $\textbf{Z}$ from the hidden representation by
computing:
\begin{dmath}
Z'=g([W'H(Z),V'H(Z)]+b')
\end{dmath}
where $W' \in \mathbb{R}^{4096 \times k}$, $V' \in \mathbb{R}^{300 \times k}$ and $b' \in \mathbb{R}^{(4096+300) \times 1}$ . Function $g$ is a sigmoid activation function. In order to learn the parameters $\theta = \{W,V, W',V',b,b'\}$ of the system, we minimize the following objective function, as detailed in \cite{chandar2016correlational}:
\begin{dmath} \label{eq:corrnet_loss}
J(\theta) = \sum_{i=1}^N L(Z_i,g(H(Z_i)))
+L(Z_i,g(H(X_i)))+L(Z_i,g(H(Y_i)))
-\lambda corr(H(X),X(Y) 
\end{dmath}

\begin{dmath}
corr(H(X),H(Y))\\
= \frac{\sum_{i=1}^N (H(X_i)-\overline{H(X)})(H(Y_i)-\overline{H(Y)}}{\sqrt{\sum_{i=1}^N(H(X_i)-\overline{H(X)})^2 \sum_{i=1}^N(H(Y_i)-\overline{H(Y)})^2}}
\end{dmath}

where $L$ is the reconstruction error, $\lambda$  is the scaling parameter used to determine the relative weight of the correlation term with respect to the reconstruction errors, $\overline{H(X)}$ is the mean vector for the hidden representations
of the first view and $\overline{H(Y)}$  is the mean vector for the hidden representations
of the second view. 

In essence, the objective function given in Equation~\ref{eq:corrnet_loss} \\
(i) minimizes the self and cross reconstruction error, and \\
(ii) maximizes the correlation between the hidden representations of the two views. This ensures that the $k$ dimensional hidden representation of text and image descriptions of the same product are very similar.

After the CorrNet is trained, we use the weights $W,V,b$ to project the input image or text query into the $k$ dimensional space of the hidden layer, which is the joint embedding space and we call it the \textit{CorrNet space}.

\subsection{Sampling from a Gaussian Mixture Model}\label{sec:sampling}
 While showing the nearest neighbors of the current user query in the \textit{CorrNet space} can provide useful recommendations, it neither captures the context of the multi-modal dialog, nor implements the exploration-exploitation paradigm. Therefore, we train the model to sample from a GMM in a way that the samples generated by the model are similar to the ground truth image recommendations for the same set of dialog rounds. We choose a Mixture of Gaussians instead of a single Gaussian with the idea that each Gaussian would be centered around certain features of the product catalog and the sampling operation would be learned in a way that the exploration-exploitation paradigm and the context are both captured by the model.
 
The model learns in a supervised framework, trying to show results that mimic the rule-based multi-modal dialog dataset results. The cosine similarity is a measure of similarity between two non-zero vectors, and since our objective is to maximize the similarity between the model generated results and ground truth results in the dataset, our objective function is the negative of the mean cosine similarity of the model generated images and the dataset images over a batch of input dialog rounds. Mathematically, the objective function is:

\begin{dmath} \label{eq:model_loss}
\mathcal{J}_m(\theta')=- \frac{1}{N}\sum_{\mathfrak{s}=1}^{N_b} \frac{1}{(N_d)^2}\sum_{j=1}^{N_d}\sum_{i=1}^{N_d} \frac {\pmb {y_i} \cdot \pmb {\hat{y_j}}}{||\pmb {y_i}|| \cdot ||\pmb {\hat{y_j}}||}
\end{dmath}
where $\mathfrak{s}$ is a sample of the batch, $N_b$ is the batch-size and $N_d$ is the number of images displayed, $\hat{y_j}$ is the $j^{th}$ model generated image and $y_i$ is the $i^{th}$ dataset image response to some query in the dialog. 
Since for each round of the dialog, the model and the ground truth dataset both contain $N_d$ images, we can form a $N_d \times N_d$ matrix of cosine similarities. For each sample, we take the mean cosine similarity from this matrix and then further consider the mean over the entire batch as our loss. As the ground truth image recommendations already take care of the exploration-exploitation paradigm, when the model learns to minimize the objective function~\ref{eq:model_loss}, it also learns to sample 
while taking care of the exploration-exploitation paradigm.

Now, sampling is a discrete operation and is non-differentiable. Hence we cannot back-propagate through a node that has a sampling operation. To overcome this problem, we use a re-parameterization technique that is commonly used in Variational Auto-encoders. A sample from a Gaussian distribution with a mean $\mu$ and covariance $\Sigma$ can be expressed as $\mathcal{z}=\mathcal{N}(\mu,\Sigma)$. Using the re-parameterization technique, we can express $\mathcal{z}$ as $\mathcal{z}=\mu+L \epsilon$, where $\epsilon = \mathcal{N}(0,\mathcal{I})$ and $\Sigma=LL^T$. So, we can now say that $\mathcal{z}$ is a function that takes parameters $(\epsilon, \mu, L)$. $\epsilon$ is an input to the network. While back-propagation, we compute partial derivatives w.r.t $\mu$ and $L$ only. The model learns the $\mu$ and $L$ for each Gaussian, as a function of the input queries and we explain this in more detail in Section~\ref{sec:model}.

Since the model learns to sample from a GMM, it should not only learn how to sample from a Gaussian distribution, but also learn how to select which Gaussian to sample from. A GMM has a pdf: $\phi_X(x)=\sum_i w_i \mathcal{N}(\mu_i, \Sigma_i;x)$. Sampling from the GMM involves the following two steps:\\
1. Sample $i$ from the categorical distribution parameterized by vector $w=(w_1,w_2, \cdots, w_d)$, with $w_i's$ such that $\sum_i w_i=1$.\\
2. Sample $x$ from normal distribution parameterized by $\mu_i$ and $\Sigma_i$.

Now, again, sampling from a categorical distribution is non-differentiable and we use the Gumbel Softmax function \cite{jang2016categorical} to overcome this problem. Sampling from a categorical distribution with class probabilities $\pi_i$ can be expressed as:
\begin{dmath}
\mathcal{z}=\texttt{one\_hot}(\underset{i}{\mathrm{argmax}}[g_i+\log \pi_i])
\end{dmath}
where $g_i$ is the gumbel noise. Since $\mathrm{argmax}$ is not differentiable, in the Gumbel Softmax technique, we use $\mathrm{softmax}$ as a continuous approximation of $\mathrm{argmax}$ as below:

\begin{dmath}
w_i = \frac{\exp((\log(\pi_i)+g_i)/\tau)}{\sum_{j=1}^C \exp((\log(\pi_j)+g_j)/\tau)} \label{eq:gumbel}
\end{dmath}

where C is the number of classes in the categorical distribution. In our case, $C$ is the number of Gaussians ($N_g$) in the GMM and is a hyper-parameter of the system. $\tau$ is called the temperature parameter that gives a control as to how closely the samples from the Gumbel-Softmax distribution would approximate the categorical distribution. When $\tau$ is very close to 0, the softmax becomes an argmax. We experiment over different values of $\tau$ and observe that $\tau = 1$ gives the best results.

The next section explains how each of the individual operations described in Section \ref{sec:method} thus far, are combined to form the complete model.

\subsection{The Complete Model}\label{sec:model}
\begin{algorithm}[htbp!]
 \SetKwInOut{Input}{Input}\SetKwInOut{Output}{Output}
\Input{\textbf{Current query $Q_t$ (text/ image click), contextual window size $N_{ws}$, Number of Gaussians $N_g$, pre-trained CorrNet weights $\{W,V,b\}$, Gumbel temperature $\tau$, Gumbel noise $g_i$, Batch size $N_b$,  Number of displayed image in each dialog round $N_d$, Learning rate $\eta$. }}
\Output{\textbf{Learned model parameters $\theta_m = \{W_{\mu}, b_{\mu}, L, W_g, b_g\}$}}
\If {$Q_t = text$}
{Find text encoding $\mathcal{T}_e$ as described in Section \ref{sec:encode}}
\ElseIf {$Q_t = image$}
{Find image encoding $\mathcal{I}_e$ as described in Section \ref{sec:encode} }
Project $Q_t$ onto the CorrNet space using $\{W,V,b\}$ to get $k$ dimensional representation $H_t$, as described in Section \ref{sec:corrNet}.\\
Compute $\mathcal{H_t} = \frac{1}{N_{ws}}\sum_{\mathcal{w}=1}^{N_{ws}-1}{H_{t-\mathcal{w}}}$.\\
The mean of each Gaussian in the GMM is computed as $\mu_i = f( \mathcal{H_t} W_{\mu i}  + b_{\mu i})$, where $W_{\mu i}$ and $b_{\mu i}$ are learned by the model and $f =\mathrm{sigmoid}$ activation function.\\
The variance of each Gaussian is computed as $\Sigma_i = L_iL_i^T$, where $L_i$ is learned by the model, but independent of $\mathcal{H_t}$.\\
Compute probability of selecting each Gaussian from the GMM as $\pi_i = g(W_g \mathcal{H_t} + b_g)$, where $W_g$ and $b_g$ are learned by the model and $g = \mathrm{softmax}$ activation function.\\
Convert sampling to a differentiable operation by the Gumbel Softmax operation as:
$w_i = \frac{\exp((\log(\pi_i)+g_i)/\tau)}{\sum_{j=1}^{N_g} \exp((\log(\pi_j)+g_j)/\tau)}$.\\
Express sampling from GMM as: $ \hat{y} = \sum_{i=1}^{N_g} w_i (\mu_i + \epsilon L_i)$, where $\epsilon \in \mathcal{N}(0,\mathcal{I})$.\\
Compute the objective function as:
$\mathcal{J}_m(\theta_m)=- \frac{1}{N_b}\sum_{\mathfrak{s}=1}^{N_b} \frac{1}{(N_d)^2}\sum_{j=1}^{N_d}\sum_{i=1}^{N_d} \frac {\pmb {y_i} \cdot \pmb {\hat{y_j}}}{||\pmb {y_i}|| \cdot ||\pmb {\hat{y_j}}||}$, where $y_i$ is the corresponding ground truth product representation.\\
Learn the model parameters $\theta_m = \{W_{\mu i}, b_{\mu i}, L_i, W_g, b_g\}$ by iteratively minimizing $\mathcal{J}_m(\theta_m)$, on the training data set.\\
Use the leaned parameters $\theta_m$ for generating results on the test data set.\\

 \caption{Algorithm for visual browsing by the agent.}
 \label{algo:modelalgo}

\end{algorithm}

\begin{figure*}[hbtp!]
  \centering
  \includegraphics[scale=0.7]{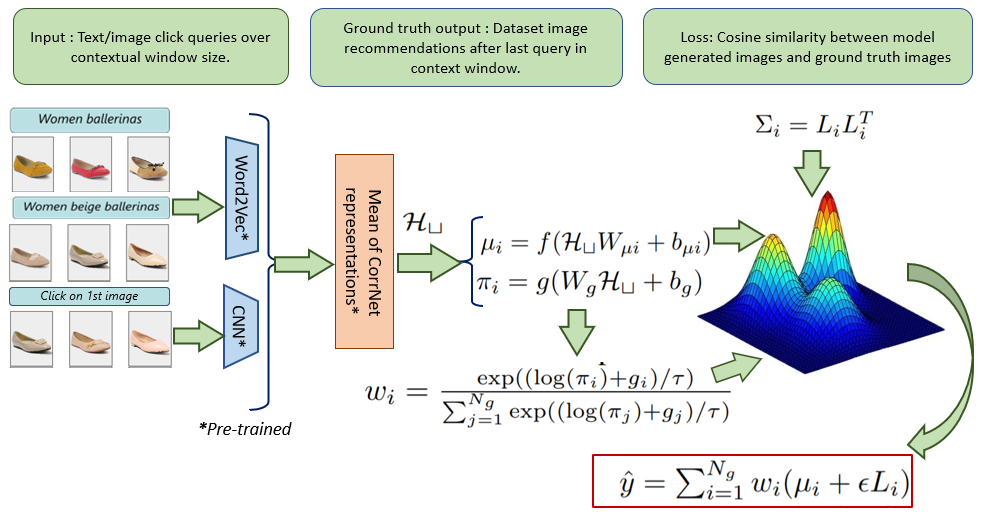}
  \caption{Schematic overview of the complete model as detailed in Algorithm \ref{algo:modelalgo}.}
  \label{fig:modelfig}
\end{figure*} 

A multi-modal dialog in the dataset can have varying number of rounds of dialog, depending on the transitions in the finite state automaton in Figure \ref{fig:automaton}. In order to keep the context in perspective, at any time instant, the model has to consider the current user query as well as the previous queries. But, how many queries should it consider to be perform optimally? It may consider all the queries in the dialog till the current query, or it may only consider the last $N_{ws}$ queries. For example, in Figure \ref{fig:modelfig}, we show 3 rounds of dialog, where the current query is a user click and the previous two queries are text queries. Thus, in this case, the contextual window-size $N_{ws} = 3$, but the model can also deal with variable window-sized context, where the algorithm allows for tuning $N_{ws}$ as a hyper-parameter.
Each encoded text and image query ($\mathcal{T}_e$ and $\mathcal{I}_e$) in $N_{ws}$ is passed through the pre-trained CorrNet to give corresponding $200 \times 1$ vectors in the CorrNet space, and the mean of these CorrNet space representations form $\mathcal{H_t}$, which is the input to subsequent parts of the network. The model takes this input  $\mathcal{H_t}$ and learns to sample from a Mixture of Gaussians. The number of Gaussians ($N_g$) is a hyper-parameter of the model. The mean of each Gaussian $\mu_i$ in the GMM, is a function of $\mathcal{H_t}$.
The probability of selecting the Gaussians in the GMM, $\pi_i$ is also a function of $\mathcal{H_t}$. This means that the model has to learn to weigh the Gaussians based on the current and the previous queries, thus capturing the context in the dialog.
The weights $w_i$ are computed from $\pi_i$ by Equation \ref{eq:gumbel} and  ultimately, sampling from the GMM can be written in the form:
$$ \hat{y} = \sum_{i=1}^{N_g} w_i (\mu_i + \epsilon L_i)$$
where $\epsilon \in \mathcal{N}(0,\mathcal{I})$.

Figure \ref{fig:modelfig} and Algorithm \ref{algo:modelalgo} show the complete model for visual browsing using Multi-Modal Dialog.

%% file: results.tex
\section {Experimental Results} \label{sec:results}
\begin{figure}[hbtp!]
  \centering
  \includegraphics[trim={3.1cm 14cm 7.6cm 2.8cm},clip, width=0.8\columnwidth]{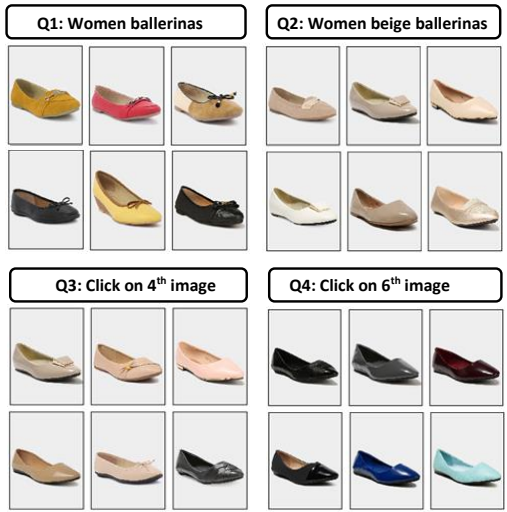}
    \caption{Four rounds of dialog and the corresponding recommendations as generated by our simulated dataset.}
    \label{fig:dialdatafig}
\end{figure}

The first contribution in this paper is the development of a multi-modal dialog dataset for visual browsing. Figure \ref{fig:dialdatafig} shows four rounds of dialog in an example session, consisting of user queries (text/image click) and our simulated dataset recommendations, generated as described in Section \ref{sec:dataset}.

\begin{figure}[hbtp!]
    \centering
    \includegraphics[trim={2.8cm 6.5in 4.5in 3.2cm},clip, width=0.49\columnwidth]{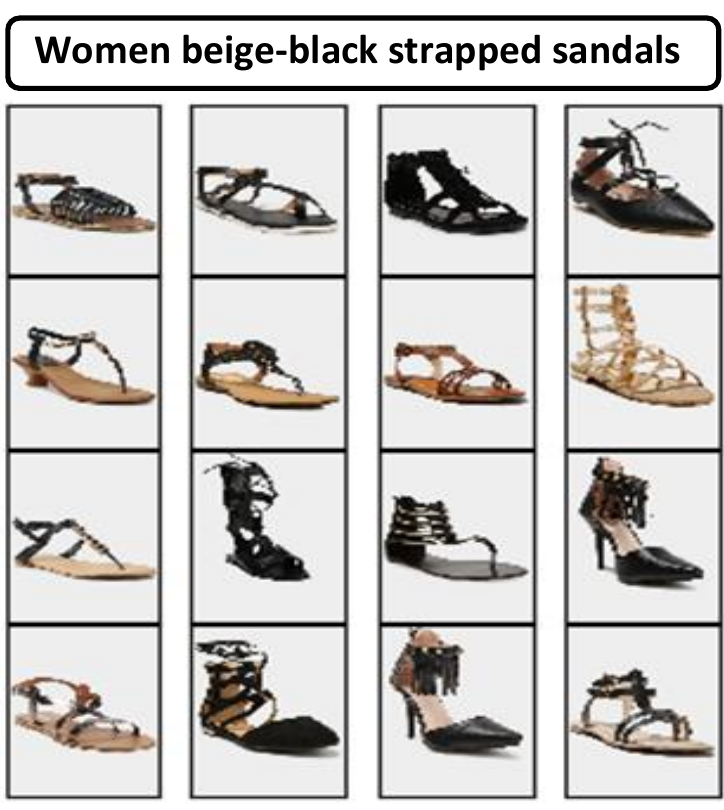}
    \hfill
    \includegraphics[trim={2.8cm 6.59in 4.5in 3.15cm},clip, width=0.49\columnwidth]{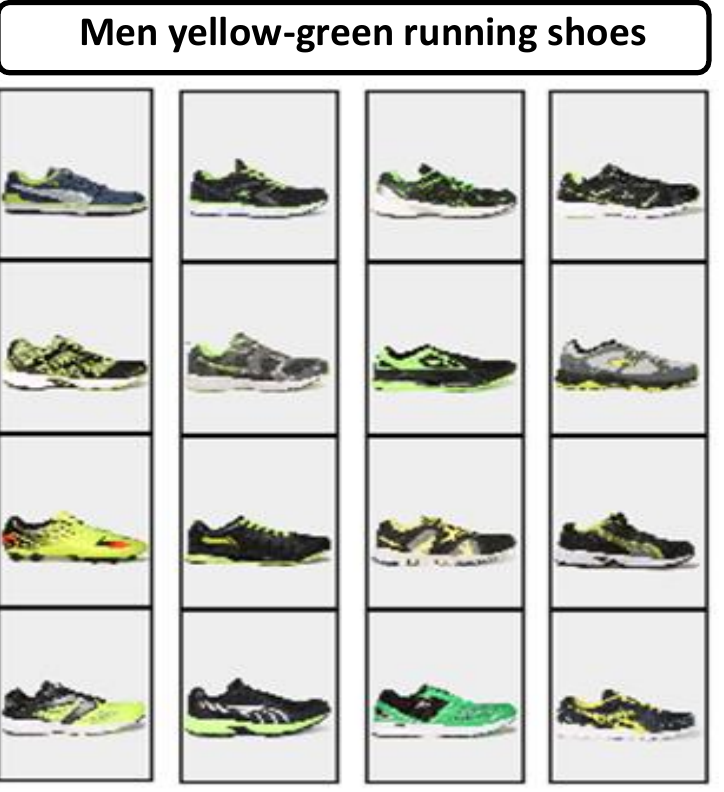}
    \caption{Images with closest CorrNet space representations to the given text representation in the CorrNet space.}
    \label{fig:corrim}
\end{figure}

Learning appropriate common representations of text and images for a product is a crucial for the complete model. After training the CorrNet model, we tested it by comparing \textit{CorrNet space} representations of text descriptions with nearest \textit{CorrNet space} image representations. By plotting the nearest image representations of a given text query, we find that the representations in the joint embedding space are very similar. Figure \ref{fig:corrim} shows this cross-referencing in the \textit{CorrNet space} for two different text representations.

\begin{figure}[ht!]
    \centering
       \includegraphics[trim={2.7cm 7.3in 1.1in 3.2cm},clip,width=\columnwidth]{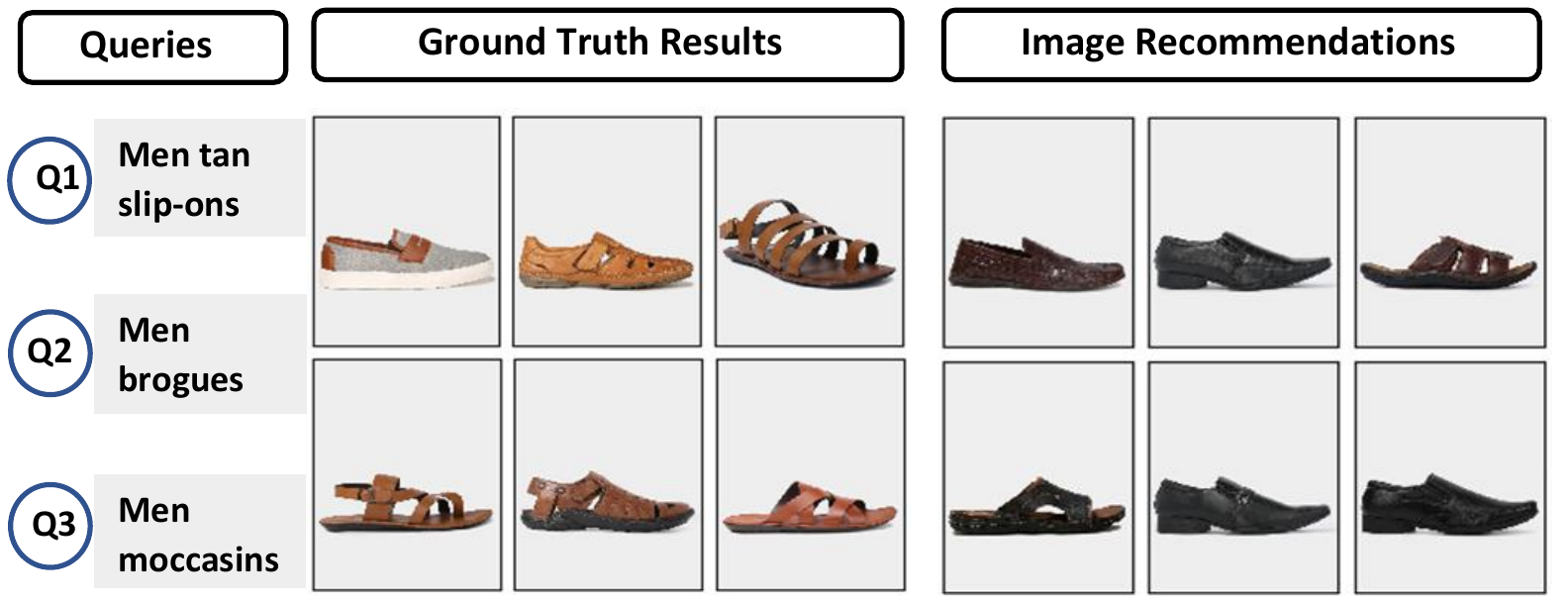}
     \includegraphics[trim={2.9cm 7.5in 1.1in 3.2cm},clip, width=\columnwidth]{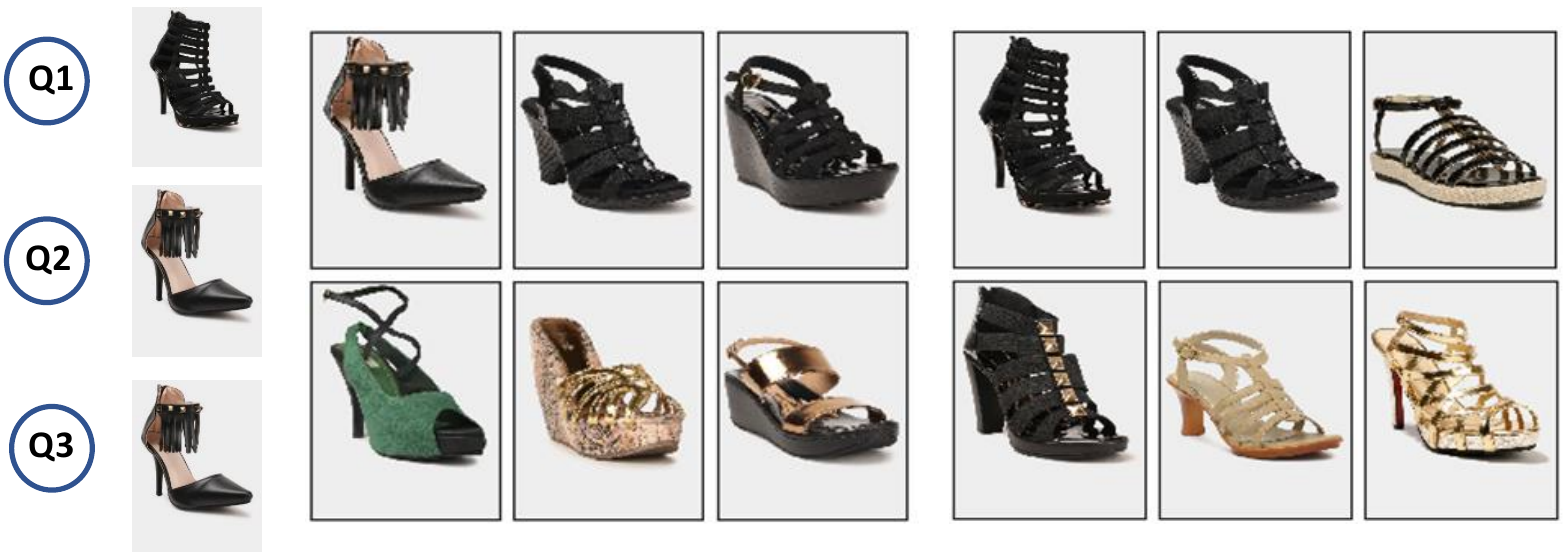}
    \includegraphics[trim={2.8cm 7.3in 1.1in 3.2cm},clip, width=\columnwidth]{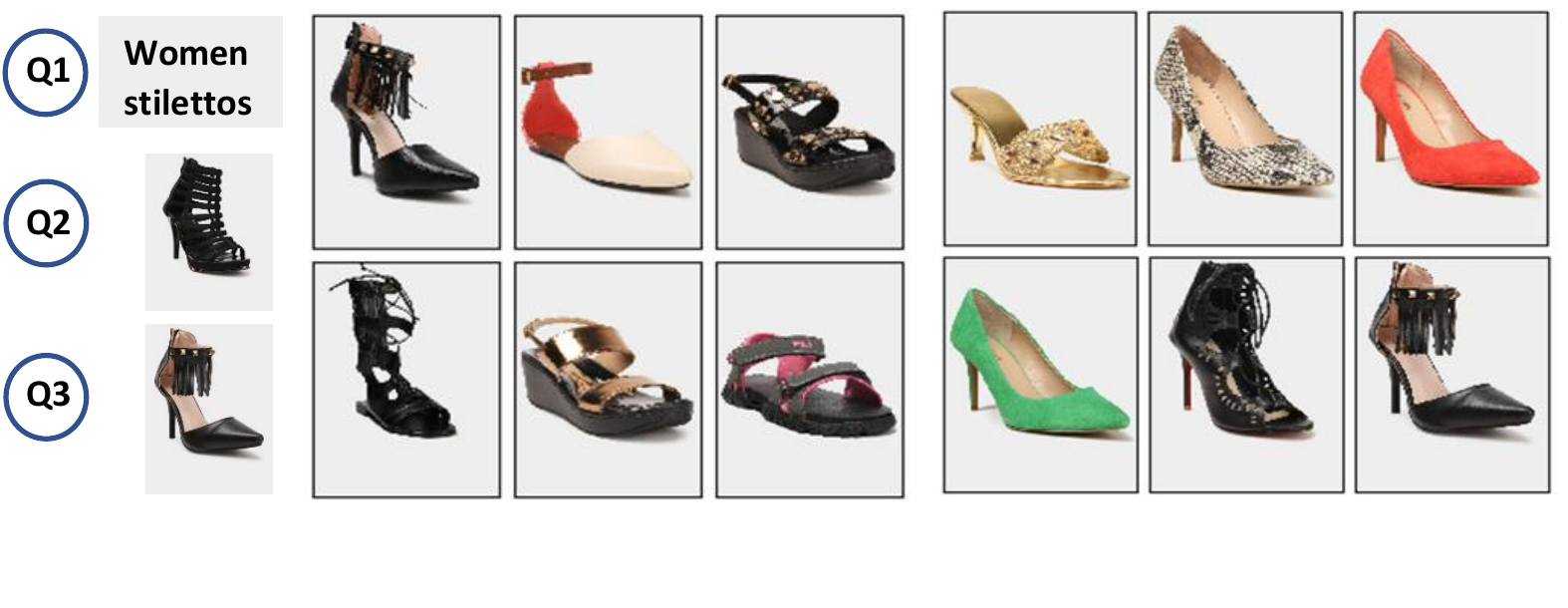}
    \vspace{-0.2in}
  \caption{Ground truth results from the dataset and the model generated recommendations, considering a contextual window size $N_{ws} = 3$. Top: Dialog session with only text queries. Middle:  Dialog session with only image click queries. Bottom : Dialog session with multi-modal queries.}
  \label{fig:resultsfig}
\end{figure}

For training the complete model, we used 70\% of the simulated dialog dataset as our training data. The hyper-parameter $k$, which is the dimension of \textit{CorrNet space} was fixed at 200 for all our experiments. We studied the performance of the algorithm with the change of hyper-parameters, the learning rate $\eta$, the temperature parameter $\tau$ in Equation \ref{eq:gumbel}, the contextual window size $N_{ws}$, and the number of Gaussians $N_g$ in the GMM. It is seen that a $\tau$ value of 1.0, an $\eta$ value of 0.1 and an $N_g$ value of 3 give optimum performance. The performance of the algorithm dwindles when the contextual window size $N_{ws}$ becomes too large, or when we have too many Gaussians $N_g$. This can be explained by the fact that the model finds it difficult to learn effectively when too much dialog history is taken into account. This can be attributed to the relatively simple method of encoding the text descriptions and the  simple averaging technique to account for the context. We believe that including a more advanced method like LSTM can help build the ``memory'' of the model more efficiently and effectively. 
Nevertheless, with simple encoding schemes and context-building, our results are quite promising. Figure \ref{fig:resultsfig} shows 3 examples of the ground truth results from our simulated dataset and the generated results by our model, when $N_{ws} = 3$. In our experiments, we achieve an average cosine similarity of 0.85 with the ground truth on the test data set.
\\

\textbf{Human Evaluation} Since, to the best of our knowledge, there exists no similar multi-modal dialog system to assist in visual browsing, it is difficult to make a direct performance comparison with any existing system. However, since the motivation of this work is to assist online customers in visual browsing, it is important to get an idea of what human users think of this system. Therefore, in order to evaluate the acceptance and performance of this multi-modal visual browsing system to human users, we conducted a preliminary user study with  13 individuals, who were undergraduate and graduate students in the age group of 23 -- 31 years.  In the user study, the participants were presented a demo version of our system where they could interact with the agent to visually browse through the shoes catalog. The participants were also presented with a system that randomly sampled products from the catalog in response to any image click or text query. The participants were free to conduct as many trial runs of either system (random sampling/ our multi-modal dialog system) as they wanted. The number of trails per participant and the number of dialog rounds per trial for each user was logged by the system. After the interaction with the system, each participant was given a questionnaire, where in addition to their age and gender, they were asked the following questions:
\begin{itemize}
\item In how many trials were you happy with the recommendations? (All of the trials, most of the trials, about half of the trials,rarely, none of the trials)
\item Do you think this kind of a recommending system can improve your online shopping experience? 
(Strongly agree, somewhat agree, neutral, somewhat disagree, strongly disagree)
\item How much do you agree, our system works better than a random baseline?
(Strongly agree, somewhat agree, neutral, somewhat disagree, strongly disagree)
\item Overall on a scale of 1 to 10 (10 being the highest), how much will you rate our system? 
\end{itemize}
 
In order to evaluate the system, we used the \textit{number of trials} and the \textit{number of dialog rounds per trial} as quantitative metrics to capture the acceptance and engagement of the users.
We evaluated the user engagement and acceptance in comparison to the random baseline system. The more the number of dialog rounds in a trial, the more the level of user engagement with the system. The number of trials by a participant is indicative of user acceptance of the system, and it suggests the user's interest as they test the system more with different starting queries. It can be taken as a proxy for an online customer coming back to re-use the system more than once.

The mean number of dialog rounds in one trial of the random sampling system was 2.8 and the mean number of rounds in one trial of our system was 4. 14.28\% of the participants conducted the random sampling experiment twice, while 46\% of the participants ran multiple trials (2 or 3) of our multi-modal dialog system. This shows that users interacted and got engaged to our system more than the random baseline and wanted to run several trials with different starting queries. The post study feedback of our system was unanimously positive. 100\% of the users strongly agreed that our system presents more meaningful results than a random recommender, and 100\% of the users agreed that such an interactive AI system for visual browsing can improve our online shopping experience. On a scale of 1 to 10 (10 being the highest), 66.7\% users rated our system as 8, and 33.3\% users gave a rating of 9. The preliminary human evaluation is promising and shows that such a multi-modal dialog system to assist visual browsing would be well-received by the online shopping community.

%% file: conclusion.tex
\section{Conclusion}

In this paper, we presented a simulated dataset of multi-modal dialogs that mimic visual browsing by a user in an online shopping environment. Using this dataset, we developed an AI agent to mimic a human agent that can remember context, understand both natural language and images, and adhere to the exploration-exploitation paradigm to assist in visual browsing. Our preliminary experiments show promising results of achieving an average cosine similarity of 0.85 between model generated results and ground truth results on the test dataset. The human evaluation study laid the foundation to the fact that such a multi-modal dialog system would be well-received by online customers, once deployed, and would be capable of engaging the customers to a greater extent than a random baseline. 

We are currently in the process of developing a larger scale dataset of multi-modal dialog using the entire range of products (all 71905 products covering apparel, footwear and accessories) in our catalog. We are also working on developing a more involved  human evaluation system on this larger dataset, by allowing users to interact with our model via a web portal, and also by comparing the performance with simple KNN based methods.

We want to develop and test more involved algorithms on this dataset including a more robust approach of ``context'' or ``memory'' using a sequential model like LSTM \cite{gers1999learning}. We also plan to extend our algorithm from the supervised learning framework to a reinforcement learning framework, where the agent can get feedback on the results. However, building a reward function for the results for this system is challenging since the goal is subjective as opposed to, say, an Atari game where we have a well defined scoring system. 

In this work, we pre-trained the CorrNet architecture to learn the joint multi-modal representation. However, this architecture cannot learn disentangled representations of the different attributes associated with a product. For example, learning disentangled representations for the important attributes like color, pattern, material etc., within a particular view and correlate them appropriately with corresponding attributes from the other view, could significantly improve the performance of visual browsing. We want to include the architecture suggested in \cite{saha2018learning}, in an end-to-end learning module, integrated with LSTM and reinforcement learning. Finally, incorporating user information and likes and dislikes can significantly improve visual browsing results, as the recommendation can be tailored to particular customers.